\documentclass[lettersize,journal]{IEEEtran}
\usepackage{amsmath}
\usepackage{amsfonts}
\usepackage{algorithmic}
\usepackage{algorithm}
\usepackage{array}
\usepackage[caption=false,font=normalsize,labelfont=sf,textfont=sf]{subfig}
\usepackage{textcomp}
\usepackage{tabularx}
\usepackage{stfloats}
\usepackage{url}
\usepackage{verbatim}
\usepackage{graphicx}
\usepackage{cite}
\usepackage{multirow}
\usepackage{multicol}

\newcolumntype{Y}{>{\centering\arraybackslash}X}


\usepackage{hyperref}

\newcommand{\mycomment}[1]{}

\begin{document}

\title{GRAP-MOT: Unsupervised Graph-based Position Weighted Person Multi-camera Multi-object Tracking in a Highly Congested Space}

\author{Marek Socha, Michał Marczyk, Aleksander Kempski, Michał Cogiel, Paweł Foszner, Radosław Zawiski, Michał Staniszewski

\thanks{This work was partly supported by European Union Funds Awarded to Blees Sp. z o. o. “Development of a system for analyzing vision data captured by public transport vehicles interior monitoring, aimed at detecting undesirable situations/behaviors and people counting (including their classification by age group) and the objects they carry” under Grant POIR.01.01.01-00-0952/20-00; and in part by the Silesian University of Technology grant for Maintaining and Developing Research Potential under Grant 02/070/BK24/0052 (MM) and 02/090/BK24/0043 (MS). Corresponding author: Michał Staniszewski.}
\thanks{Marek Socha is with the Department of Data Science and Engineering, Silesian University of Technology, 44-100 Gliwice, Poland.}
\thanks{Michał Marczyk is with the Department of Data Science and Engineering, Silesian University of Technology, 44-100 Gliwice, Poland, and also with the Yale Cancer Center, Yale School of Medicine, New Haven, CT 06510 USA.}
\thanks{Aleksander Kempski, Paweł Foszner, and Michał Staniszewski are with the Department of Computer Graphics, Vision and Digital Systems, Silesian University of Technology, 44-100 Gliwice, Poland (e-mail: mstaniszewski@polsl.pl).}
\thanks{Aleksander Kempski is with the Department of Systems Biology and Engineering, Silesian University of Technology, 44-100 Gliwice, Poland.}
\thanks{Radosław Zawiski is with the Department of Automatic Control and Robotics, Silesian University of Technology, 44-100 Gliwice, Poland.}
\thanks{Michał Cogiel is with Blees Sp. z o. o., 44-100 Gliwice, Poland and QSystems.pro Sp. z o. o., 41-907 Bytom, Poland}
\thanks{Manuscript received XX; revised XX}
}

\markboth{Journal of \LaTeX\ Class Files,~Vol.~14, No.~8, August~2021}%
{Shell \MakeLowercase{\textit{et al.}}: A Sample Article Using IEEEtran.cls for IEEE Journals}

\IEEEpubid{0000--0000/00\$00.00~\copyright~2021 IEEE}

\maketitle

\begin{abstract}
GRAP-MOT is a new approach for solving the person MOT problem dedicated to videos of closed areas with overlapping multi-camera views, where person occlusion frequently occurs. Our novel graph-weighted solution updates a person's identification label online based on tracks and the person's characteristic features. To find the best solution, we deeply investigated all elements of the MOT process, including feature extraction, tracking, and community search. Furthermore, GRAP-MOT is equipped with a person’s position estimation module, which gives additional key information to the MOT method, ensuring better results than methods without position data. We tested GRAP-MOT on recordings acquired in a closed-area model and on publicly available real datasets that fulfil the requirement of a highly congested space, showing the superiority of our proposition. Finally, we analyzed existing metrics used to compare MOT algorithms and concluded that IDF1 is more adequate than MOTA in such comparisons. We made our \href{https://gitlab.qsystems.pro/publicly/grap-mot}{https://gitlab.qsystems.pro/publicly/grap-mot} along with the acquired \href{https://doi.org/10.5281/zenodo.14526116}{https://doi.org/10.5281/zenodo.14526116} publicly available.
\end{abstract} 


\section{Introduction}
Monitoring human activity across multiple overlapping camera views in closed environments presents unique challenges and opportunities. While a single camera may suffer from occlusions or limited visibility, overlapping fields of view allow for improved robustness and continuity in tracking. Multi-camera multi-object tracking (MOT) enables the integration of fragmented observations into coherent trajectories, which is essential for consistent scene understanding and downstream analytics in applications such as security monitoring, flow analysis, or resource optimization. Our work addresses this problem by proposing a method that operates effectively in such constrained settings without relying on prior identity information or full-body visibility.

Thanks to the increased amount of captured video recordings, the algorithms for person detection, recognition, or tracking are currently developing rapidly. Person detection refers to denoting an individual with a rectangle \cite{2001_first_obj}, while tracking refers to assigning a unique identifier (ID) to a rectangle while ensuring that between subsequent video frames the identifier remains constant \cite{2006_tracking_survey}. This approach can be extended to the multi-camera multi-object tracking, where more than one camera is incorporated in the tracking process. As the image sources are different, to match objects properly across cameras, more effort has to be put in. Due to occlusions, non-overlapping fields of view (FOV), changes in angle, and lighting, this problem remains a challenge. A related task is person re-identification, which focuses on retrieving instances of the same person across different cameras, typically under non-overlapping FOV conditions. However, unlike re-identification, MOT does not assume the prior availability of gallery images and often deals with overlapping camera views. Additionally, MOT requires not only maintaining consistent identity labels across time within a single camera view but also ensuring correct identity matching across different cameras.

In the given work, we introduce a novel method named GRAP-MOT for tracking many persons in a closed area using at least 2 different cameras with overlapping views, placed to cover all areas of the space from different angles. The multi-camera multi-object tracking (MOT) task in such an environment is problematic due to numerous occlusions, changes in viewing angle, and strongly differing person detection sizes. Using an internal dataset where the scene is usually highly condensed, has limited time for data acquisition, and the persons tend to stand still during records, we provided a deep analysis of each element of the MOT problem to choose the final solution. Thus, we proposed a complete system capable of tracking individual persons on a single camera and tracking multiple persons when many cameras are used. We tested our solution on a specially prepared mock-up simulating a closed space with a different number of people, as well as on available datasets meeting the assumed requirements. Our source code\footnote{https://gitlab.qsystems.pro/publicly/grap-mot} along with the acquired data\footnote{https://doi.org/10.5281/zenodo.14526116} are publicly available. 
The main contributions of this paper are summarized as follows:

\begin{itemize}
  \item In contrast to other MOT methods, GRAP-MOT relies on the strengthening of the tracklets (i.e. a sequence of short detections) association over time rather than assuming immediate proper track matching. It makes convergence longer but gives higher robustness of detections.
  \item GRAP-MOT relies only on constant observation of the weighted graph and ongoing updating of ID numbers using acquired tracks, information about a person's position, and their characteristic features. No other data are needed.
  \item Based on the information from the graph, the method uniquely analyzes possible connections and creates object-tracking groups based on their similarities.
  \item A module for estimating people's positions, not available elsewhere, is a key element of multi-camera multi-object tracking. 
  \item GRAP-MOT requires little to no supervision. The only trained model is the one for feature extraction, which could also be acquired from public repositories.
  \item Additionally, we conducted an important analysis of comparative metrics for assessing the effectiveness of MOT, proposing the IDF1 against the MOTA, which frequently appears in other works. 
\end{itemize}


\subsection{Review of existing tracking methods}
Person detection and tracking methods are well-developed in the literature (Table \ref{tab:papers}). Early works in the area of person detection consisted of either simple shape delineation methods based on active contours \cite{1997_geo_active_contours, 1998_snakes_shapes} or face detection by image features \cite{2001_simple_face_features}. Since 2015, there has been a rapid rise in interest regarding detection methods driven by the development of new deep learning models \cite{2015_deep_learning}. The most popular detection networks are R-CNN \cite{2015_rcnn}, Faster R-CNN \cite{2015_faster_rcnn}, Mask R-CNN \cite{2017_mask_rcnn}, and YOLO \cite{2016_yolo} method with its multiple versions. Tracking aids detection by allowing one to follow objects across video frames, thus bridging the gaps between detections and frames. In 2006 the tracking methods were split into multiple categories \cite{2006_object_tr_rev} including point-based tracking \cite{1960_kalman, 1990_point_track} where a point in space represents the tracked object, kernel-based methods \cite{2003_kernel_based, 2001_kernel_based_svm} referring to object shape and motion and silhouette tracking \cite{1998_sithoulet, 1993_hausdorff} where the object contour and image features are the matching factor. Currently, the most common subjects for detection and tracking are rectangular detections, so-called bounding boxes, for which point-based tracking methods are particularly popular. Usually, those methods are based on the Kalman Filter \cite{1960_kalman} due to its simplicity but high efficiency. Methods like POSE \cite{1996_POSE_kalman} or SORT \cite{2016_sort} utilize the filter to predict the next position solely based on the current position and motion. Later, hybrid systems emerged, connecting the point-based and silhouette categories. Methods like DeepSORT \cite{2017_deepsort, 2018_deepsort}, ByteTRACK \cite{2022_bytetrack}, or MOTDT \cite{2018_motdt}, which are currently regarded as the state-of-the-art, use deep learning methods to define object features and employ the information from the Kalman filter to support their decision.

Multi-camera multi-object tracking (MOT) ensures that across multiple cameras, detection of the same person or object shares the same identifier \cite{2013_reid_review}. These methods can be divided according to the location, the analyzed object, the distribution of the objects, the number of cameras, the degree of camera overlap, and the timing of the recording. In recent years, we have witnessed an increase in the popularity of the MOT methods development, mostly due to the AI city challenge \cite{2020_ai_city_chall, 2021_ai_city_chall}, resulting in many works centered around the tracking of motor vehicles. The deep learning-based solutions are predominant in this area. For example, researchers used them to extract features of the objects and employed GPS location to re-identify and track cars \cite{2022_cars_mc_mt_luna, 2022_cars_mc_mt_reId}. Common practice is to employ the cars' trajectories to better match tracks between cameras \cite{2020_cars_tracklets_matching_mc_ct}. Found tracks can be matched to the reference camera \cite{2020_cars_single_camera_to_mcameras_match} or analyzed simultaneously using, for example, graph methods \cite{2022_cars_by_graphs_mc_mt_reId, 2020_cars_tracklets_matching_mc_ct, 2019_cars_graphb_trajectory_mc_mt_non_overlapping}.

The problem of tracking people in a highly congested closed space, where occlusion frequently appears and there is higher variation in detection sizes, is much harder than tracking in an open area. There is a limited number of works covering MOT in these types of scenes. For example, MOT in the operating room uses skeletal pose estimation with tracking to compensate for the lack of colorful clothes and visible faces \cite{2022_operation_room_mc_mp_reid}. The method for person tracking in the warehouse employs elaborate image processing to deal with the sudden changes in the lighting due to the large windows \cite{2020_warehouse_shanghai_mc_mp_reid, 2023_warehouse_motion_features}. To deal with the person's tracking in the shop, the authors used the approach revolving around the density maps since the camera was placed on the ceiling \cite{2020_peron_in_shop_mc_mp}. The same authors proposed later a similar approach, but with the addition of a trajectory showing the improvement over previous work \cite{2020_person_reind_with_trajectory_mc_mp}. Most recent approaches tend to use the spatial and temporal graph approach, where the spatial graph matches tracks on the nth frame across different views and temporal associates objects between frames. For example, the DyGLIP method \cite{2021_person_mc_mp_non_overlapping} uses a dynamic graph network with attention to match spatial information and then match it with temporal tracks. The ReST separates these tasks into two, separately trained graphs; the first graph combines spatial information, and the second graph has the task of combining tracks in time between successive frames \cite{2023_rest}.

The problem of person re-identification, which focuses on comparing each probe image to a predefined set of gallery images and selecting the most similar match, is also well described. In \cite{nguyen2024tackling}, person re-identification (Person ReID) is understood as the task of matching individual people from images collected from multiple non-overlapping cameras. The process in practice involves comparing a query image with an image gallery, and this paper is a review of methods based on deep learning models to deal with domain shifts, including clothing changes, in Lifelong ReID scenarios. The re-identification task is defined in the same way in the case of the work \cite{zhang2024view}, but is considered in the context of multiple heterogeneous (airborne and ground-based) cameras. Through this, a view-difference arises, and in response, the authors propose a View-decoupled Transformer (VDT) model, based on the Transformer architecture (more precisely ViT-Base), which aims to separate view-related features from features crucial for identity identification. In this paper, the authors used two datasets containing both ground and aerial views, their proprietary synthetic CARGO (Civic AeRial-GrOund) dataset and the actual AG-ReID dataset. The work \cite{qin2024noisy} focuses on the problem of text-image person re-identification (TIReID), the understanding of which is different from previous work. Re-identification is defined as the task of finding an image of a person in a large image gallery based on a textual description. The main challenge addressed in the paper is the problem of noisy correspondence (NC) in training data, and the authors propose a Robust Dual Embedding (RDE) method that uses OpenAI's CLIP-B/16 pre-trained model as the underlying encoders for learning robust visuo-semantic associations. The work \cite{li2024all} considers the problem of re-identifying persons (Person ReID), relative to image galleries based on different types of input data such as image, text, or sketch. To make this possible and to increase generalisation, the authors propose an AIO model that uses a frozen, pre-trained base model based on the Vision Transformer (ViT) architecture as a shared feature encoder. The work uses both real and synthetic collections. Due to the lack of a large amount of annotated data, sketches are sometimes generated from RGB images. The work performs evaluations on independent collections such as Market1501 and CUHK-PEDES, for example. The paper \cite{xu2024distribution} also considers re-identification as a probe image search in a gallery, but considers it in the context of Lifelong Person Re-identification - LReID. It proposes a Distribution-aware Knowledge Prototyping (DKP) method, the premise of which is not to store learning data, bypassing the privacy issues and computational costs associated with memory-based methods. Instead, it creates prototypes to store this information, creating a distribution for each sample, taking into account its individual variability. The method is trained on Market1501, DukeMTMC-reID, CUHK-SYSU, MSMT17-V2 and CUHK03 image data, and is tested on non-dependent sets. In the work \cite{xu2022rank}, re-identification is also referred to as an image gallery search process. The main innovation of the work is to propose a new loss function, called Differentiable Retrieval-Sort Loss (DRSL), to optimise the feature distribution, which is used with typical re-ID model architectures such as ResNet-50 used as the core of the network. The work performs evaluations on sets such as Market1501, CUHK03, and MSMT17, for example. The paper \cite{ding2024text} introduces the concept of re-identification as Text-to-Image Vehicle Re-Identification, which is understood as searching for a target vehicle by matching a text description with images in a photo gallery. The context is urban surveillance systems using multiple non-overlapping cameras. To reduce the gap between modalities (text and image), the authors propose a Multi-scale multi-view Cross-modal Alignment Network (MCANet), which uses ResNet-50 for vision and BERT with text convolution modules for text as the core. The work \cite{wu2024intermediary} focuses on the re-identification of people visible on non-overlapping cameras based on image galleries. However, in addition to using RGB images for this purpose, the authors additionally use depth images of images. To reduce the significant discrepancies between RGB modality and depth, the authors propose an Intermediary-Generated Bridge Network (IBN), a network using a ResNet50-based architecture as the core, complemented by a multi-modal transformer and circle contrast learning module. In this work, the authors use real data from the RobotPKU, BIWI, and SYSU-MM01 collections.

The idea of using graphs for classic re-identification was used previously in the literature. However, since all of these methods assume that there is a pre-defined reference database of known objects, they cannot be used in the MOT task defined above. In \cite{shen2018deep}, person re-identification is formulated as the task of locating a target individual in an image gallery by comparing a probe image against all gallery images. While most existing methods rely solely on probe-to-gallery (P2G) similarities, the proposed Deep Group-shuffling Random Walk Network incorporates both P2G and gallery-to-gallery (G2G) similarities into an end-to-end learning framework. The model is based on a Siamese CNN architecture using ResNet-50 as the backbone, and its effectiveness is demonstrated on standard multi-camera datasets: Market-1501, CUHK03, and DukeMTMC. Similarly, \cite{shen2018person} defines re-identification as matching a probe image within a gallery. The proposed Similarity-Guided Graph Neural Network (SGGNN) integrates both P2G and G2G relationships to refine similarity estimation during both training and inference. The method also builds on a Siamese CNN framework and is evaluated on Market-1501, CUHK03, and DukeMTMC. In \cite{li2023camera}, the re-identification task focuses on vehicles across non-overlapping camera networks. The approach uses ResNet-50 for initial feature extraction and introduces a Camera Topology Graph Convolutional Network (CT-GCN), which explicitly models spatial and directional relationships between cameras. The graph is constructed with four hierarchical levels: system-wide (global connection), position (spatial proximity), orientation (directional alignment), and individual (self-loop). This structure allows the model to learn more discriminative, camera-independent features. The method in \cite{bao2019masked} addresses person re-identification across disjoint camera views using a Masked Graph Attention Network (MGAT). It employs ResNet-50 for feature extraction and leverages a graph attention mechanism to model global relationships among all gallery images. This enables more effective refinement of features for probe-gallery matching. The approach is trained and tested on iLIDS-VID, PRID2011, MARS, and Market-1501 datasets. Finally, \cite{wang2024vision} tackles vehicle re-identification by defining it as an unguided search task across multiple camera views. The method uses OpenAI’s CLIP-B/16 model to extract joint visual and textual features and constructs a graph (VLCGT) to model relationships among training samples. This graph is incorporated into the learning process to improve similarity estimation and enhance feature discriminability.

\begin{table}[!t]
    \caption{Summarization of papers similar to our problem. P: people/person MOT, CS: closed space, HD: High density, OV: Overlapping views, CA: Code is available\label{tab:papers}}
    \centering
    \begin{tabular}{|@{}||l||l||c||c||c||c||c||@{}|}
        \hline
        Ref & Year & P & CS & HD & OV & CA \\
        \hline
        \cite{2023_rest} & 2023 & \checkmark & - & \checkmark & \checkmark & \checkmark \\
        \hline
        \cite{2023_cctv_reid} & 2023 & \checkmark  & - & - & -  & -  \\
        \hline
        \cite{2023_warehouse_motion_features} & 2023 & \checkmark  & \checkmark & - & \checkmark  & \href{https://github.com/yuntaeJ/SCIT-MCMT-Tracking}{\checkmark}  \\
        \hline
        \cite{2023_non_overlapping} & 2023 & \checkmark  & - & - & -  & -  \\
        \hline
        \cite{2023_fisheye_camera_network} & 2023 & \checkmark  & \checkmark & \checkmark & \checkmark  & -  \\
        \hline
        \cite{2022_operation_room_mc_mp_reid} & 2022 & \checkmark  & \checkmark & - & \checkmark  & -  \\
        \hline
        \cite{2022_cars_mc_mt_reId} & 2022 & \checkmark & - & \checkmark & - & - \\
        \hline
        \cite{2022_cars_mc_mt_luna} & 2022 & - & - & - & - & \checkmark\\
        \hline
        \cite{2022_cars_by_graphs_mc_mt_reId} & 2022 & - & - & - & - & \checkmark \\
        \hline
        \cite{2021_centroids_reid} & 2021 & \checkmark & - & - & - & \href{https://github.com/mikwieczorek/centroids-reid}{\checkmark} \\
        \hline
        \cite{2021_person_mc_mp_non_overlapping} & 2021 & \checkmark & - & - & \checkmark & \checkmark \\
        \hline
        \cite{2020_person_reind_with_trajectory_mc_mp} & 2020 & \checkmark & - & - & \checkmark & \href{https://github.com/GehenHe/TRACTA}{\checkmark} \\
        \hline
        \cite{2020_peron_in_shop_mc_mp} & 2020 & \checkmark & \checkmark & \checkmark & - &  -\\
        \hline
        \cite{2020_warehouse_shanghai_mc_mp_reid} & 2020 & \checkmark & \checkmark & - & \checkmark  & -\\
        \hline
        \cite{2020_cars_tracklets_matching_mc_ct} & 2020 & - & - & - & \checkmark & - \\
        \hline
        \cite{2020_cars_single_camera_to_mcameras_match} & 2020 & - & - & - & \checkmark & - \\
        \hline
        \cite{2020_unsupervised_reid} & 2020 & \checkmark & - & - & - & \href{https://github.com/DengpanFu/LUPerson}{\checkmark} \\
        \hline
        \cite{2019_temporal_reid} & 2019 & \checkmark & - & - & - & \href{https://github.com/Wanggcong/Spatial-Temporal-Re-identification}{\checkmark} \\
        \hline
        \cite{2018_pose_1} & 2018 & \checkmark & - & - & \checkmark & \href{https://github.com/yxgeee/FD-GAN}{\checkmark} \\
        \hline
        \cite{2018_pose_2} & 2018 & \checkmark & - & - & - & - \\
        \hline
    \end{tabular}
\end{table}

\section{Materials and Methods}

\subsection{Internal Dataset}

\begin{figure*}[!t]
    \centering
    \subfloat[]{\includegraphics[width=0.3\textwidth]{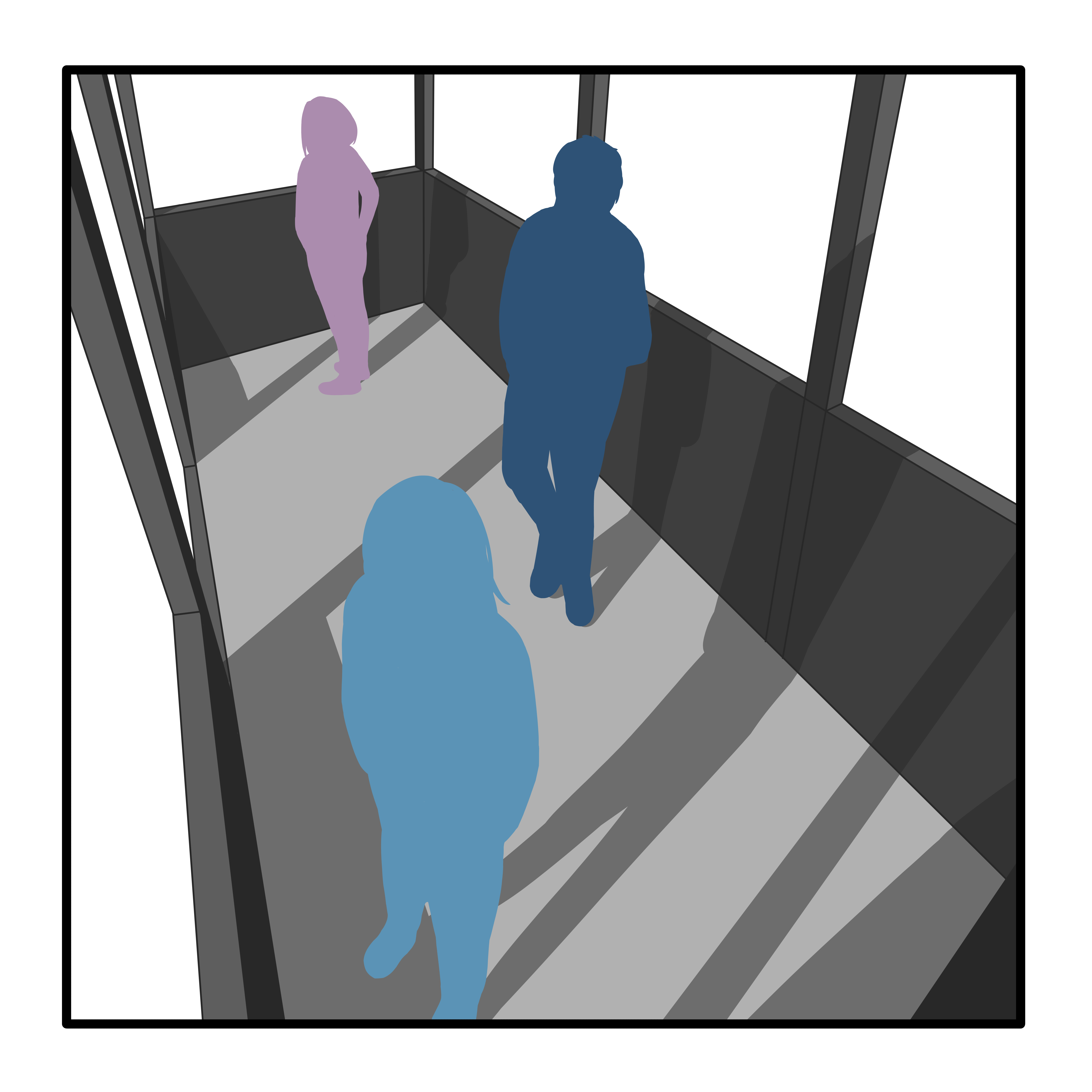}%
    \label{fig:left_camera}}
    \hfil
    \subfloat[]{\includegraphics[width=0.3\textwidth]{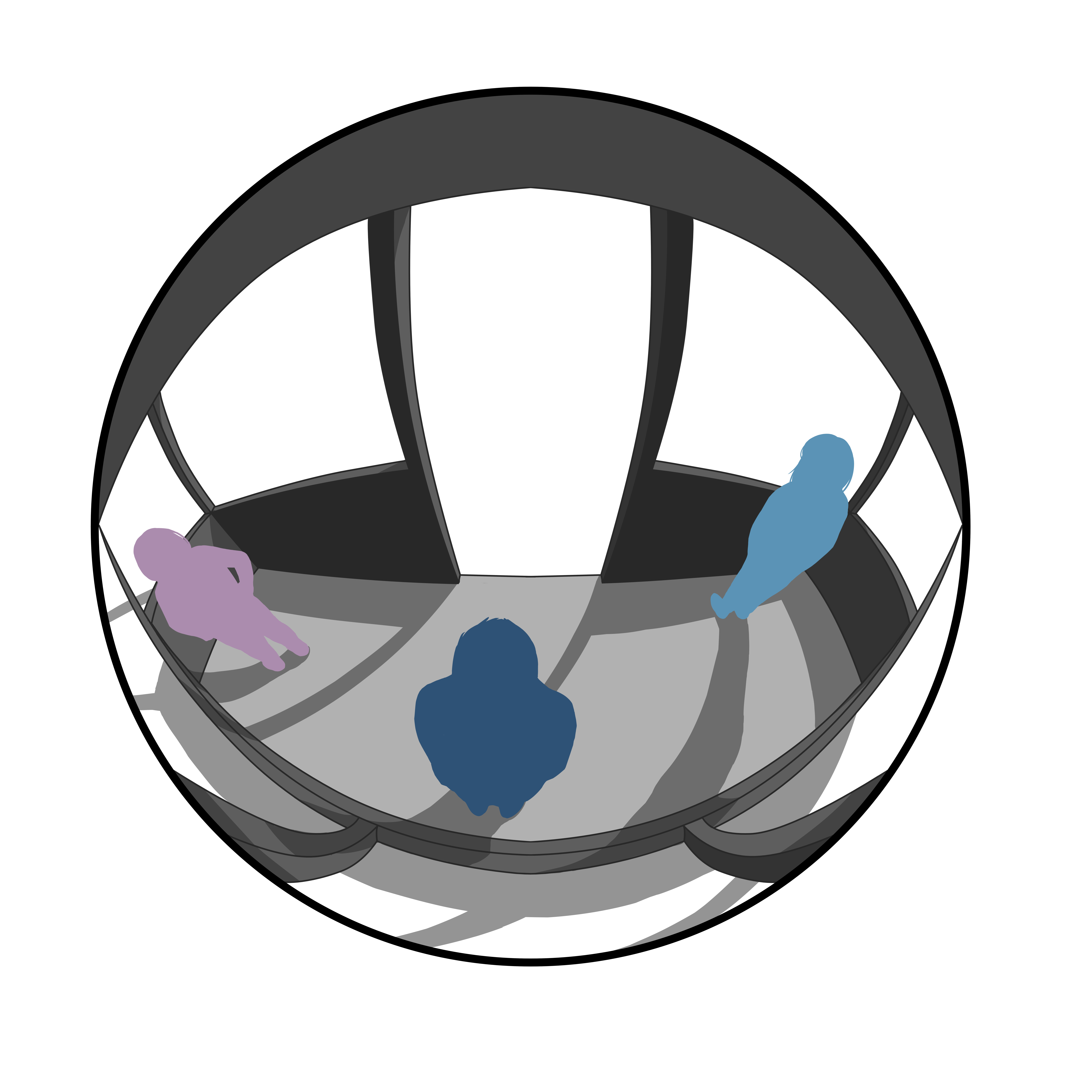}%
    \label{fig:fish_eye_camera}}
    \hfil
    \subfloat[]{\includegraphics[width=0.3\textwidth]{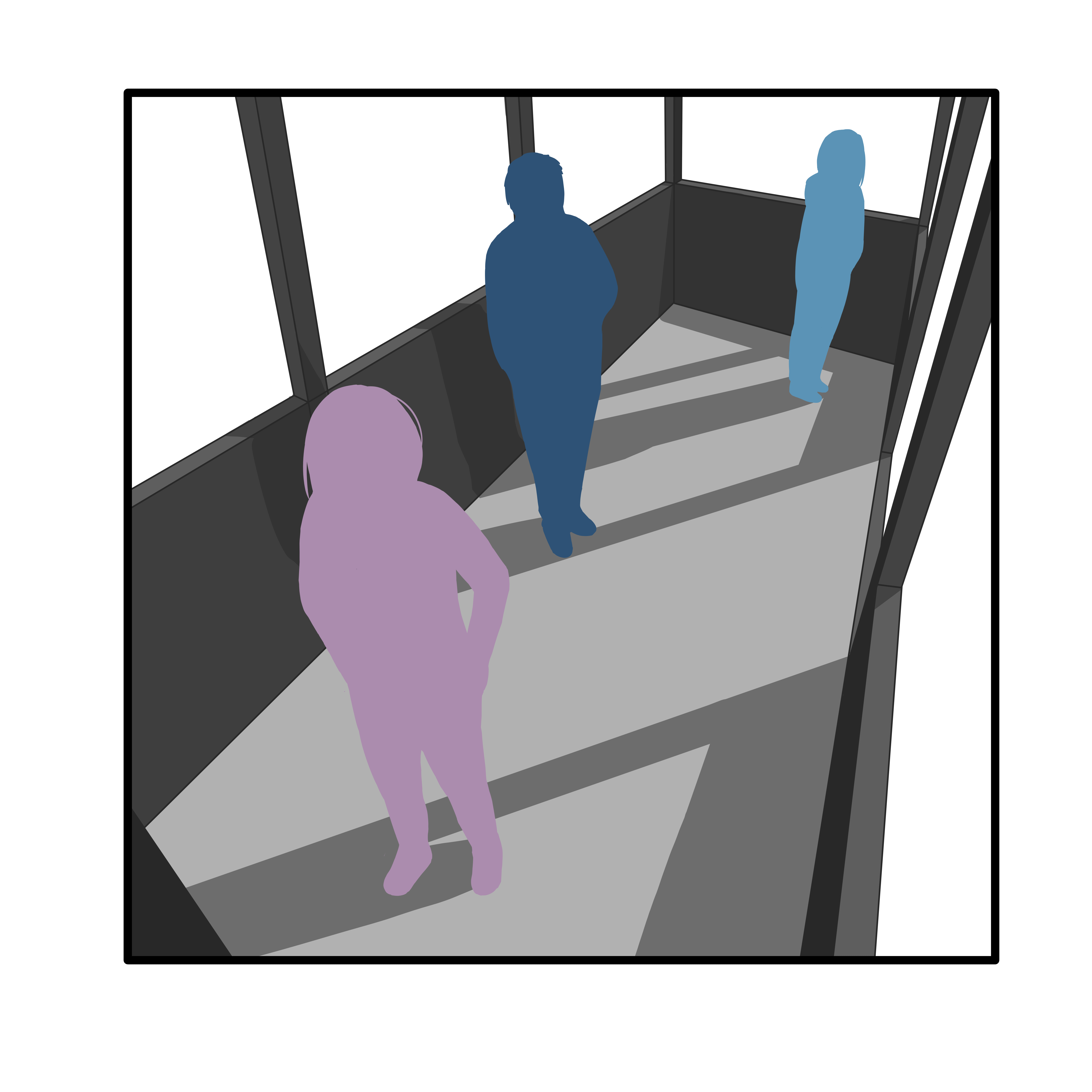}%
    \label{fig:right_camera}}
\caption{Iconographic visualization of camera distribution on the closed-area model frame. The exemplary person's multi-object tracking on different camera views is included in different colors, where left/right (a and c) are typical views and (b) in the middle is a fish-eye type view.}
\label{fig:cameras_distribution}
\end{figure*}

A closed-area model was constructed in the laboratory to acquire an internal benchmark dataset. Three cameras were installed, including two standard cameras in adjacent corners of the model (Figure \ref{fig:left_camera}, Figure \ref{fig:right_camera}) and one fish-eye camera (Figure \ref{fig:fish_eye_camera}) in the center near the entrance. Such camera placement allows the entire area inside the model to be visible without leaving blind spots. A total of 14 scenes were recorded, capturing the behaviour of numerous people, ranging from 2 to 15 people per recording. Each scene was created in a closed-area model and comprises 10 video recordings lasting about 10 seconds, and 252 frames on average. Using a detection model based on the YOLOv7 \cite{2023_YOLOv7} architecture, persons' heads were found and surrounded by bounding boxes to support bounding box labeling for tracking. Head detections were curated by manually correcting missing detections, adjusting the bounding box area to the persons' heads, and removing incorrect detections. After data curation, detections were labeled to match across all cameras. 

\subsection{External datasets}
The CAMPUS \cite{2016_campus_anot_ecp} public dataset was used as an external validation set. The head detections and annotations made in the study of Yuan Xu \textit{et al.} \cite{2016_campus_anot_ecp, 2017_anot_ecp} were used. CAMPUS dataset \cite{2016_campus_anot_ecp} contains 4 recordings with a partially overlapping field-of-view (FOV) taken from 4 cameras, each with a resolution of 1980x1020, and recorded at a 30 fps frame rate. The Auditorium subset comprises footage captured by two cameras placed at the entrance and two within the auditorium, amounting to about 5,000 frames. The Garden1 and Garden2 subsets feature recordings from four cameras in the park, each with overlapping fields of view, containing 3,000 and 6,000 frames, respectively. The Parkinglot subset includes videos from four cameras situated in various spots in the parking area, also with overlapping fields of view, comprising 6,500 frames.

\mycomment{\begin{figure}[h]
    \centering
    \includegraphics[width=1\linewidth]{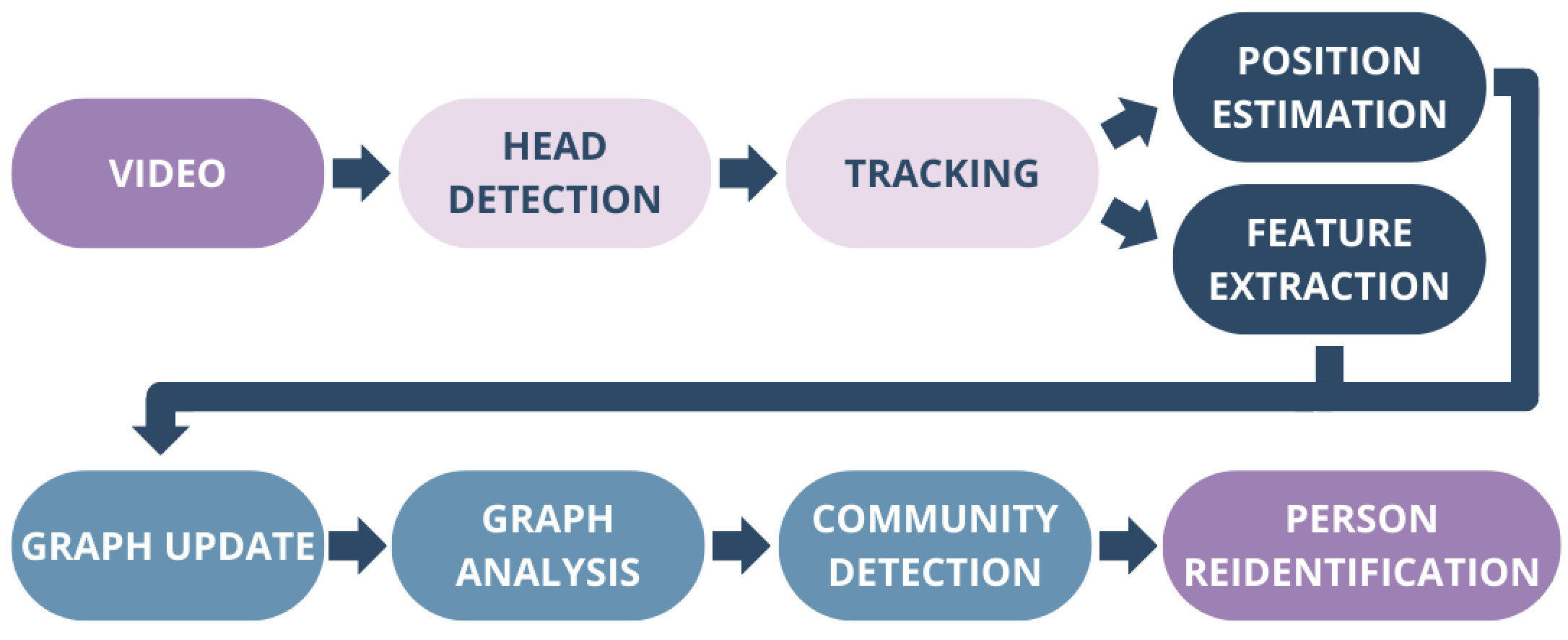}
    \caption{The general block diagram of the GRAP-MOT pipeline.}
    \label{fig:block-diagram}
\end{figure}}

\begin{figure*}[!t]
    \centering
    \includegraphics[width=0.98\textwidth]{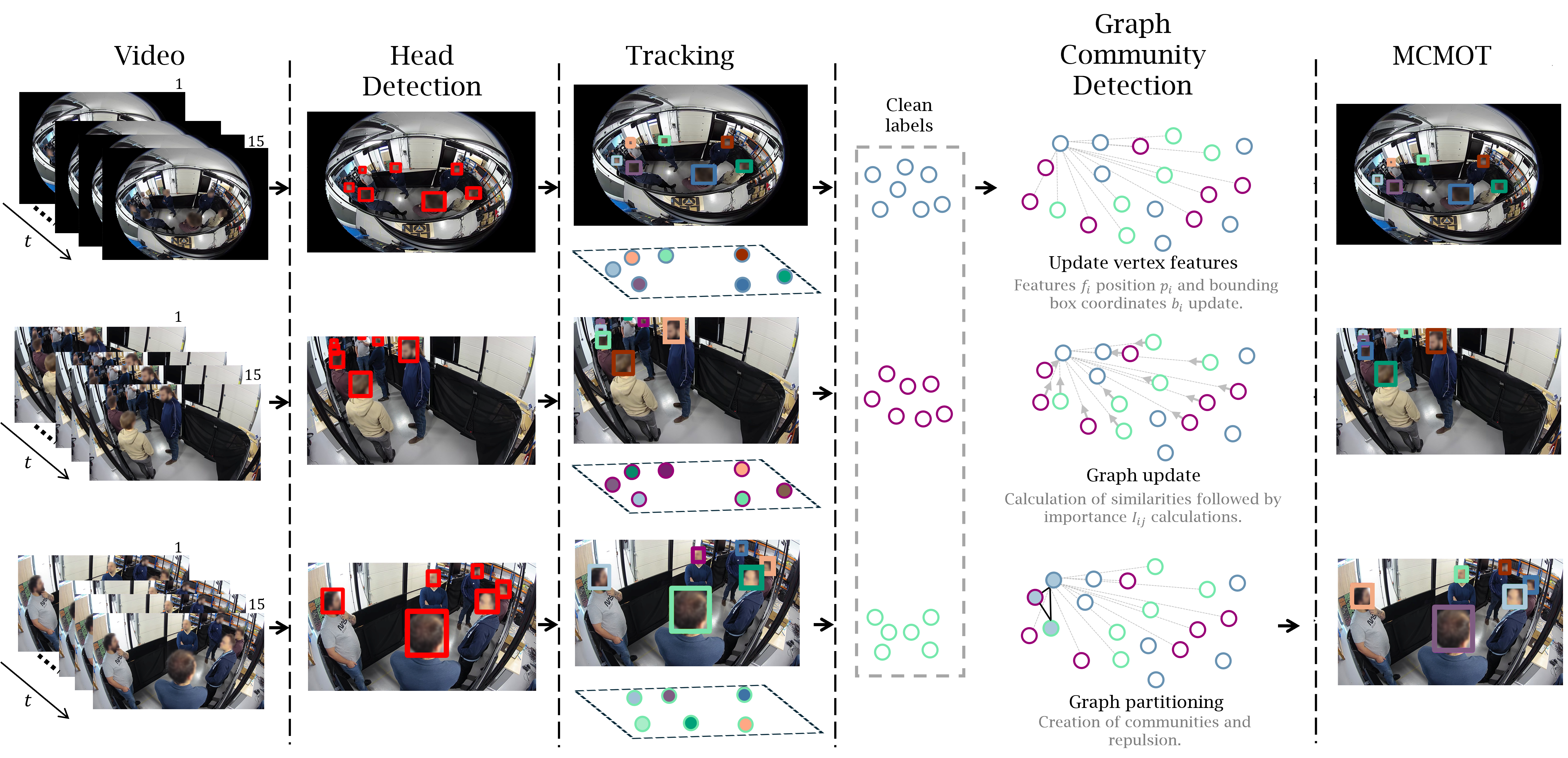}
    \caption{Overview of the GRAP-MOT Multi-Camera Multi-Object Tracking Pipeline. Video frames from multiple cameras are processed sequentially. A head detection module outputs bounding box coordinates for each detected individual. Within each camera view, a tracking algorithm assigns temporary identity labels to detections, forming short-term trajectories (tracklets). A position estimation module projects these tracklets onto a common spatial grid. All tracklets are then represented as nodes in a graph, where edges connect tracklets originating from different cameras. An importance coefficient is computed for each edge to quantify the likelihood of cross-camera identity correspondence. Tracklets are clustered across cameras into groups, with group sizes limited by the total number of cameras. Finally, unified identity labels are assigned within each group, yielding consistent tracking across all camera views.}
    \label{fig:visual-block-diagram}
\end{figure*}

\subsection{Head detection method}

YOLO (You Only Look Once) version 7 \cite{2023_YOLOv7} was trained to detect human heads on images from a closed-area model. According to our previous work \cite{2024_position}, head detections are more robust to occlusions and more stable in comparison to full-body detections. From the available models, the YOLOv7-X was chosen as a compromise between precision and evaluation time for the head detection task. In training, default parameters were used together with a batch size equal to 8 and the number of epochs equal to 300. Two datasets were used to train the model: (i) the CrowdHuman benchmark dataset \cite{shao2018crowdhuman}; (ii) our internal dataset of images taken in the closed-area model  \cite{2024_position}. The CrowdHuman consists of 470 thousand human unique instances with an average of 23 persons per image and different levels of occlusions. From this database 15,000, 4,370, and 5,000 images were used for training, validation, and testing, respectively. From our internal dataset, 543 images were used for model training and 61 for model validation.

\subsection{Tracking methods}
To track people within a single camera the following methods were evaluated: SORT \cite{2017_deepsort}, DeepSORT \cite{2018_deepsort}, FairMOT \cite{2021_fairmot}, and ByteTrack \cite{2022_bytetrack}. These methods follow similar steps of analysis. Given the bounding box from the object detection method, the first step is to predict the next position of the tracking object. Secondly, there is a data association phase where predicted positions are matched with similar previous positions, creating a tracklet, i.e. a sequence of short detections (in contrast to track, which is a whole trajectory of an object). Finally, there is a track management phase where the tracks are updated, added, or deleted.

Every tested tracking method follows the presented workflow but each provides some novel tracking solution. The oldest method SORT is a point-tracking algorithm based on a Kalman filter which predicts the object's future location using only the current state (position and velocity). The association phase uses the Hungarian algorithm to match tracklets with predicted positions based on the bounding boxes' Intersection over Union (IoU). The DeepSORT method was created to enhance the association step by adding the appearance features extracted by the deep learning model tailored by the user to match the tracked object. Matching is done by combining IoU and cosine similarities between the appearance features. FairMOT attempts to unify position detection and feature extraction into one network, making it efficient but less flexible. ByteTrack introduced a novel way of dealing with low-confidence detections where such detections are matched during the association phase with the existing tracklets making them more likely to appear in the future. Additionally, low-confidence detections are used to prevent loss of the tracklet during occlusions.

\subsection{Feature Extraction}
Different convolutional neural networks were tested for feature extraction from head images, namely OSNet \cite{2019_osnet}, ONet-AIN \cite{2021_osnet_ain}, and ResNet50 \cite{2015_resnet}, which were trained for the MOT task on the Market1501 \cite{2015_market1501} and DukeMTMC \cite{2016_duke_mtmc} collections with the torchreid repository \cite{torchreid}. The training set included all images from the collections mentioned above (see description of datasets). The networks were trained with 100 epochs at max, with input sizes set to 256x180 and images augmented using random flip and random crop methods. Head images were usually smaller than the desired size so they were up-scaled using nearest-neighbours interpolation method.

\subsection{Position estimation}

The XGBOOST \cite{chen2016xgboost} model was used to estimate a person's position, which was originally used in \cite{2024_position} to predict the X and Y coordinates of the person's position in a bus, and also for the decision if each person is inside or outside of the bus. The most important parameters were tuned using the Bayesian optimization method \cite{snoek2012practical}: (i) eta - step size shrinkage used in the model update (range 0.001-0.5); (ii) max depth - maximum depth of a tree (range 1-20); (iii) gamma - minimum loss reduction required to make a further partition on a leaf vertice of the tree (range 0-0.1); (iv) colsample bytree - subsample ratio of columns when constructing each tree (range 0.4-1); (v) min child weight - minimum sum of instance weight needed in a child (range 0.1-10); (vi) subsample - subsample ratio of the training instances that occur once in every boosting iteration (range 0.5-1); (vii) lambda - L2 regularization term on weights (range 0-10); (viii) alpha - L1 regularization term on weights (range 0-10). Selection of the best parameters was done using 10-fold cross-validation. The position estimation was calculated only for the internal dataset. In the case of the external datasets, the tracklets did not have this attribute.

\subsection{Graph update}

The cameras are denoted as $c$, and tracklets are denoted as $T$. The number of cameras is constant and equal to $C$. The number of tracklets per camera may not be equal. Let us denote the total number of tracklets as $M$ and treat them as a single set, despite their different camera origins, for simplicity. The method assumes that the tracklets contain extracted object features ($f$), object position estimation ($p$), detection bounding box coordinates ($b$), and the camera of the origin ($c$) (Equation \ref{eq_tracklet_definition}). 

\begin{equation}
    T_i = \{f_i, p_i, b_i, c\} \;\;\;\; c \in [1..C] \;\;\;\; i \in [1..M]
    \label{eq_tracklet_definition}
\end{equation}

To combine tracklets from multiple cameras, a graph defined by the set of vertices $V$ (Equation \ref{eq_vertices_definition}) and the set of edges $E$ (Equation \ref{eq_edges_definition}) is constructed. In vertices, tracklet information is stored. Only vertices with tracklets that originate from different cameras are connected by edges (tracklets from the same camera are not connected). Thus, edges represent the inter-camera relation between the tracklets.

\begin{equation}
    V=\{v_1, v_2, ..., v_{M}\}
    \label{eq_vertices_definition}
\end{equation}

\begin{equation}
    E=\{\{v_i, v_j\} \mid i \neq j \land c_i \neq c_j\} \;\;\;\;  i,j \in [1..M]
    \label{eq_edges_definition}
\end{equation}

After creating the graph, the cosine distance between the object features and between the position estimates is calculated for each edge. The resulting distances are normalized to the range from 0 to 1 using the sigmoid ($\sigma$) function and converted to similarities for features $s_{ij}^f$ and for positions $s_{ij}^p$. Additionally, in each edge, occurrences ($o_{ij}$) are initialized with the value 1. The occurrence parameter counts the number of video frames in which two vertices connected by the edge were labelled as the same person.

\begin{equation}
     s_{ij}^f = 1-\sigma \left( \dfrac {f_i \cdot f_j} 
     {\left\| f_i\right\| _{2}\left\| f_j\right\|_{2}} \right)
     \;\;\;\; 
     i \neq j \land c_i \neq c_j  \;\;\;\;  i,j \in [1..M]
     \label{eq_features_distance}
\end{equation}

\begin{equation}
     s_{ij}^p = 1- \sigma \left( \dfrac {p_i \cdot p_j} 
     {\left\| p_i\right\| _{2}\left\| p_j\right\| _{2}} \right)
     \;\;\;\; 
     i \neq j \land c_i \neq c_j \;\;\;\;  i,j \in [1..M]
     \label{eq_position_distance}
\end{equation}

To quantify the inter-camera relation between tracklets, for each edge, the importance value $I$ is calculated. The edge importance value is updated at each frame and is defined by the equation (Equation \ref{eq_importance_definition}). The larger the $I$ on the edge between the vertices, the more probable they correspond to the same person. Division by three keeps the $I$ update value in the range of 0 and 1, where auxiliary frame-related index $t$ is used to indicate consecutive updates of the $I$ parameter.

\begin{equation}
    I_{ij} = I_{ij}^{t-1} +  \frac{1}{3} \left( s_{ij}^p + s_{ij}^f + \frac{o_{ij}}{\sum_{l=1}^{M-1}\sum_{k=l+1}^{M} o_{lk}} \right)
    \label{eq_importance_definition}
\end{equation}

If the position estimates are unavailable, a different form of the equation to calculate edge importance (Equation \ref{eq_importance_coords_definition}) is used.

Let $K=\{k_1, k_2, ..., k_{M_k}\}$ and $L=\{l_1, l_2, ..., l_{M_l}\}$ represent vertices (tracklets) of the graph from the two different cameras.  For each vertex $k_i$ (tracklet), the cosine distance between the bounding box coordinates of $k_i$ and other vertices from the same camera is calculated. This creates an intra-camera neighbour vector $INV$ (Equation \ref{eq_inv}). The operation is repeated for each camera. The goal is to compare the values of the vector $INV$ between vertices from different cameras. Thus, each vector $INV$ of $K$ camera vertices is compared to each vector $INV$ of $L$ camera vertices (Equation \ref{eq_irm_comp}), giving $IRM$ matrix. Finally, the bounding box relation coefficient $r$, which reflects neighbourhood similarities of the tracklets between cameras, is calculated as the arithmetic mean of the previous value (i.e. initial value or previous frame value) and the sum of $IRM$.

\begin{equation}
    INV_{k}=\dfrac {b_{k_i} \cdot b_{k_j}} 
     {\left\| b_{k_i}\right\| _{2}\left\| b_{k_j}\right\| _{2}} \;\;\;\;  i \in [1..M_k]
    \label{eq_inv}
\end{equation}

\begin{equation}
    INV_{k}=\left[ INV_{k_1}, INV_{k_2}, ..., INV_{M_k} \right]
    \label{eq_inv}
\end{equation}

\begin{equation}
    IRM_{k_il_j} = (1-|INV_{k_i}-INV'_{l_j}|)^{2} \;\;\;\;  i \in [1..M_k] \;\;\;\;  j \in [1..M_l]
    \label{eq_irm_comp}
\end{equation}

\begin{equation}
    r_{ij}^t=\frac{r_{ij}^{t-1} + \sum IRM_{ij}}{2}
\end{equation}

The calculated bounding box relation coefficient $r$ is then used to calculate the edge importance using the alternative equation that relies more on the similarity between features $s^f$ and has $r$ as the support for the decision (Equation \ref{eq_importance_coords_definition}). For this purpose, the smoothed occurrences $q_{ij}$ coefficient is defined (Equation \ref{eq_occurence_smoothing_definition}) where $\alpha$ is the smoothing factor set to 50 for the analyzed problem. Division by $1+o_{ij}$ scales the $I$ update value to the range 0 and 1, assuring numerical boundness.

\begin{equation}
    q_{ij} = 1 - \exp(\frac{-1}{\alpha}  o_{ij})
    \label{eq_occurence_smoothing_definition}
\end{equation}

In the proposed importance equation, bounding box features are the main anchor. The bounding box relation coefficient comes online after the anchor was established, meaning the occurrence value was updated for a given connection. Only then is $q$ different from 0.

\begin{equation}
    I_{ij}^t = I_{ij}^{t-1} +  \frac{ r_{ij} q_{ij} + s_{ij}^f} {1 + o_{ij}}
    \label{eq_importance_coords_definition}
\end{equation}

\subsection{Graph partitioning}

\begin{figure}[h]
    \centering
    \includegraphics[width=1\linewidth]{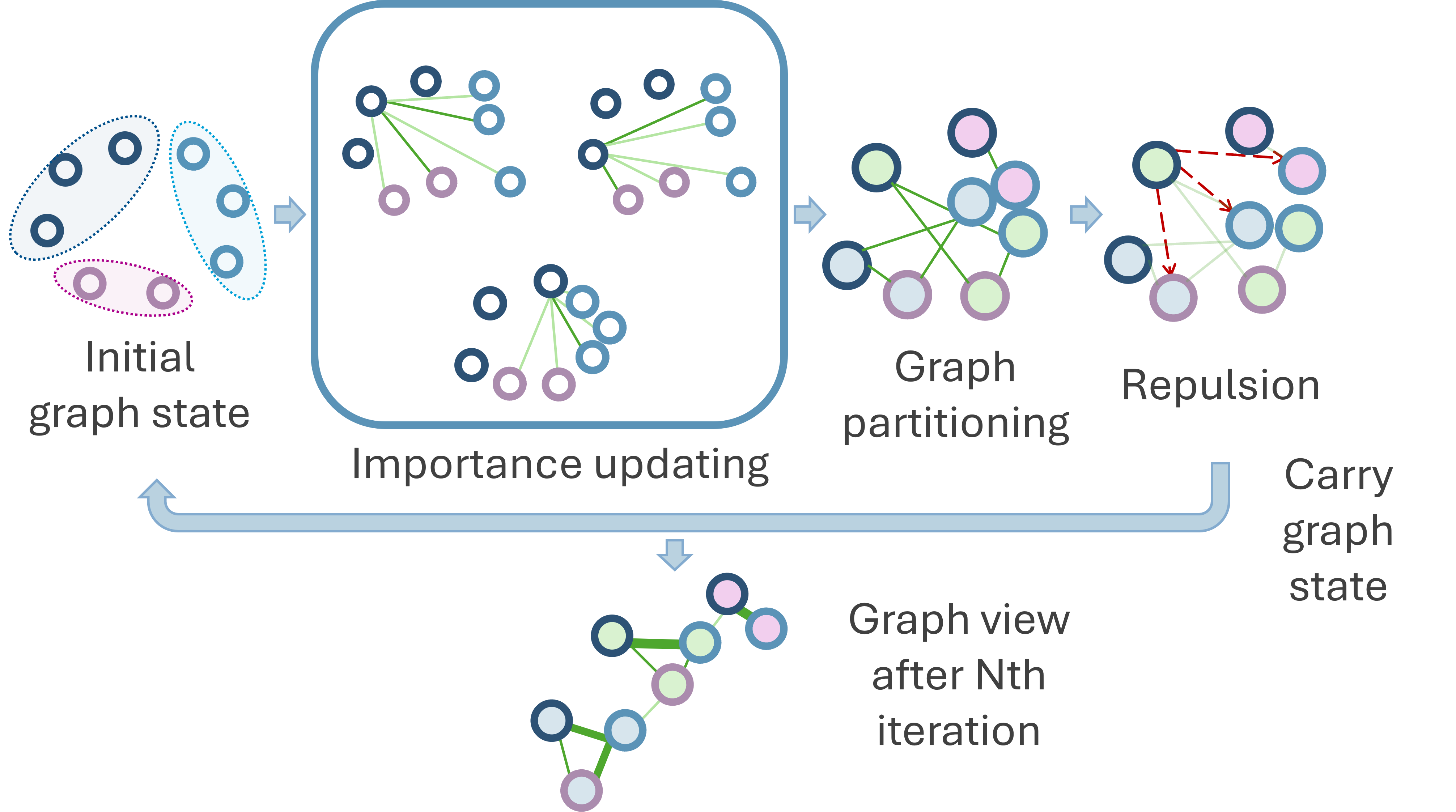}
    \caption{Overview of graph update, partitioning and repulsion processes.}
    \label{fig:methods-graph-operations}
\end{figure}

To organize the vertices in groups, communities were created. Communities are groups of points, in this case, vertices, which commonly interact with each other. The interaction between the vertices is simulated in the presented method by the Importance value stored in the edges of the graph. For this purpose, the greedy modularity communities detection algorithm was used \cite{2004_newman_greedy_mod_det_com}.  The algorithm was modified to limit the maximum number of vertices in the society to the number of cameras used in the MOT task. The modularity measure is used to quantify the graph community partitioning quality. Given the list of communities, it rates them based on the number of connections inside the community. The equation also has a weighted variation (Equation \ref{eq:modularity}) where a given edge feature present between the vertices in the community replaces several connections. In both cases, classic and weighted versions, modularity is defined as:

\begin{equation}
    Q = \frac{1}{2m} \sum_{ij} \left( A_{ij} - \gamma\frac{k_ik_j}{2m}\right) \delta(c_i,c_j)
    \label{eq:modularity}
\end{equation}

where $m$ is the sum of the edge weights, $A$ is the adjacency matrix, $\gamma$ is the resolution parameter, $k$ is the weighted degree of $i$ (or $j$), and $\delta(c_i,c_j)$ is a Kronecker delta (if $i$ and $j$ are in same community 1, if are in different 0). The $\gamma$ resolution parameter determines whether intra-group or inter-group relationships are more important. 

After the communities were created, they were validated by checking whether each community had tracklets from the same cameras. In such an event, vertices in the validated community are repulsed from each other by subtracting the Repulsion coefficient $e$ (Equation \ref{"eq_repulsion_coeff"}) of the current frame from the stored Importance value, making it less probable for vertices to join each other communities in the future. The repulsion coefficient is defined for the vertex as the sum of the neighboring edge occurrences divided by the number of neighbors ($N_n$).

\begin{equation}
    e = \frac{\sum_{i=1}^{N_n}o_i}{N_n}
    \label{"eq_repulsion_coeff"}
\end{equation}

\begin{equation}
    I_{ij} = I_{ij}^{t-1} - e
    \label{"eq_new_importance_after_rep"}
\end{equation}

\subsection{Evaluation metrics}
There are two groups of commonly used metrics for multi-camera multiple-object tasks, the CLEAR-MOT group \cite{2008_clear_mot, 2016_clear_mot} and the ID group \cite{2016_ID_multiobject_metrics}. The former focuses mainly on assessing tracking and detection quality, and the latter on assessing match quality. From these groups, the two most important metrics for the MOT task were selected. MOTA (Multiple Object Tracker Accuracy) measures overall tracking and detection accuracy, considering missed detections, mismatches, identity switches, and false positives. A high MOTA value indicates good tracking quality. IDF1 measures the quality of identifier assignment for tracks. It takes into consideration both precision and recall of identities across frames. High IDF1 indicates that appropriate identifiers are maintained throughout the frame sequence.

The pymotmetrics package \cite{2023_pymot} was used to calculate the listed metrics. To calculate them for multi-camera multiple-object tasks, each camera was assigned an identifier from 0 to $N_c$ (number of cameras). Detection results were added sequentially frame by frame. Since the frame number has to be unique, even if the results come from different cameras, the frame number was multiplied by 1000, and the camera ID was added. In this way, MOTA and IDF1 could be calculated for a multi-camera problem. 

\subsection{Statistical analysis}

Multiple tracking methods, feature extraction, and community detection were tested during experiments. To compare the difference between the results of different methods, the Kruskal-Wallis test was used. In all tests, the statistical significance level was set to $\alpha=0.05$ . If there was enough evidence to state the inequality of medians, the test was followed by Nemenyi post-hoc test. If during the test of method 1 against method 2, the p-value was smaller than the significance level and the median result of method 1 was greater than the median of method 2, method 1 was considered better.

\section{Results}

In our experiments, the baseline configuration combined DeepSORT for tracking, ResNet50 for feature extraction, and greedy modularity optimization for community detection. We then evaluated the impact of changing each module individually and of disabling the position estimation module. Median IDF1 and MOTA metrics were computed per dataset to compare configurations. The optimal combination was subsequently benchmarked against other methods on the external dataset.

\subsection{Performance of the tracking methods in a congested space}

At first, we analyzed the performance of different object tracking methods on a single camera for the MOT problem in the highly congested space. IDF1 and MOTA metrics were calculated for 140 recordings, and metrics were aggregated by median value based on the number of people present in the scene (Supplementary Table 1, Supplementary Figure 1). In Table \ref{tab:tracking_idf1_mota}, we present the median, mean, and standard deviation of the results for each tracking method. We found that the obtained IDF1 metrics are statistically different between methods (p-value=0.0245, $\alpha=0.05$). To find the best-performing method, we confronted post hoc Nemenyi test results with the median of the sets' scores. According to the analysis, the best tracking method is SORT, since it is better than all other methods (relative to DeepSORT p-value=2.624e-02, FairMOT p-value=4.796e-14, and Byte Track p-value=0.0186). The second best proved to be Byte Track, which was better than DeepSORT and Fair MOT (relative to DeepSORT p-value=0.0262, Fair MOT p-value=7.132e-11). The DeepSORT method, which was a default method for the experiments, placed third, proving to be better than Fair MOT (relative to Fair MOT p-value=3.991e-04).

\begin{table}[]
\centering
\caption{Tracking methods' performance comparison on the internal dataset. The median, mean and standard deviation of the create group metrics are shown in the table.}
\label{tab:tracking_idf1_mota}
\begin{tabular}{|c|c|ccc|}
\hline
\multirow{2}{*}{\textbf{Tracker}}   & \multirow{2}{*}{\textbf{Metric}} & \multicolumn{3}{c|}{\textbf{Aggregation}}                                                     \\ \cline{3-5} 
                                    &                                  & \multicolumn{1}{c|}{\textbf{Median}} & \multicolumn{1}{c|}{\textbf{Mean}}   & \textbf{STD}    \\ \hline
\multirow{2}{*}{\textbf{DeepSORT}}  & \textbf{IDF1}                    & \multicolumn{1}{c|}{79.279}          & \multicolumn{1}{c|}{77.608}          & \textbf{14.701} \\ \cline{2-5} 
                                    & \textbf{MOTA}                    & \multicolumn{1}{c|}{63.689}          & \multicolumn{1}{c|}{64.202}          & 21.339          \\ \hline
\multirow{2}{*}{\textbf{SORT}}      & \textbf{IDF1}                    & \multicolumn{1}{c|}{\textbf{81.492}} & \multicolumn{1}{c|}{\textbf{79.975}} & 15.115          \\ \cline{2-5} 
                                    & \textbf{MOTA}                    & \multicolumn{1}{c|}{72.269}          & \multicolumn{1}{c|}{72.950}          & 18.938          \\ \hline
\multirow{2}{*}{\textbf{FairMOT}}   & \textbf{IDF1}                    & \multicolumn{1}{c|}{75.799}          & \multicolumn{1}{c|}{74.021}          & 17.189          \\ \cline{2-5} 
                                    & \textbf{MOTA}                    & \multicolumn{1}{c|}{70.506}          & \multicolumn{1}{c|}{70.110}          & 20.228          \\ \hline
\multirow{2}{*}{\textbf{ByteTrack}} & \textbf{IDF1}                    & \multicolumn{1}{c|}{80.969}          & \multicolumn{1}{c|}{78.980}          & 15.074          \\ \cline{2-5} 
                                    & \textbf{MOTA}                    & \multicolumn{1}{c|}{70.037}          & \multicolumn{1}{c|}{71.414}          & 20.285          \\ \hline
\end{tabular}
\end{table}

\subsection{Searching for the best model architectures for a person's head image}

After selecting the most effective single-camera tracking method, we tested the performance of different neural network models for feature extraction from a person's head bounding box. IDF1 and MOTA were aggregated by a median concerning their specific set (Supplementary Table 2, Supplementary Figure 2). In Table \ref{tab:nn_models_idf1_mota}, we present the median, mean, and standard deviation of the results for each neural network model used for feature extraction. We found statistical differences between the results of different models. After the post hoc Nemenyi test and median result analysis, the best-performing network proved to be the ResNet50 model (relative to OSNet x1 p-value=4.4409e-14, OSNet x0.5 p-value=6.05071e-14, OSNet AIN x1 p-value=5.773e-14 and OSNet IBN p-value=7.549e-13). The differences in results of other model architectures were not statistically significant (p-values$>$0.05). Thus, there is no difference in the median of the quantitative variable across models aside from the ResNet50. The other model architectures were placed equally in last place.

\begin{table}[]
\centering
\caption{Neural network models' performance comparison on the internal dataset. The median, mean and standard deviation of the create group metrics are shown in the table.}
\label{tab:nn_models_idf1_mota}
\begin{tabular}{|c|c|ccc|}
\hline
\multirow{2}{*}{\textbf{Model}}        & \multirow{2}{*}{\textbf{Metric}} & \multicolumn{3}{c|}{\textbf{Agreggation}}                                                     \\ \cline{3-5} 
                                       &                                  & \multicolumn{1}{c|}{\textbf{Median}} & \multicolumn{1}{c|}{\textbf{Mean}}   & \textbf{STD}    \\ \hline
\multirow{2}{*}{\textbf{ResNet50}}     & \textbf{IDF1}                    & \multicolumn{1}{c|}{\textbf{79.279}} & \multicolumn{1}{c|}{\textbf{77.608}} & \textbf{14.701} \\ \cline{2-5} 
                                       & \textbf{MOTA}                    & \multicolumn{1}{c|}{63.689}          & \multicolumn{1}{c|}{64.202}          & 21.339          \\ \hline
\multirow{2}{*}{\textbf{OSNetx1}}      & \textbf{IDF1}                    & \multicolumn{1}{c|}{63.004}          & \multicolumn{1}{c|}{64.768}          & 22.696          \\ \cline{2-5} 
                                       & \textbf{MOTA}                    & \multicolumn{1}{c|}{52.388}          & \multicolumn{1}{c|}{55.848}          & 25.253          \\ \hline
\multirow{2}{*}{\textbf{OSNet x0.5}}   & \textbf{IDF1}                    & \multicolumn{1}{c|}{62.245}          & \multicolumn{1}{c|}{64.271}          & 22.895          \\ \cline{2-5} 
                                       & \textbf{MOTA}                    & \multicolumn{1}{c|}{47.439}          & \multicolumn{1}{c|}{55.507}          & 25.827          \\ \hline
\multirow{2}{*}{\textbf{OSNet AIN x1}} & \textbf{IDF1}                    & \multicolumn{1}{c|}{63.135}          & \multicolumn{1}{c|}{63.926}          & 22.015          \\ \cline{2-5} 
                                       & \textbf{MOTA}                    & \multicolumn{1}{c|}{50.678}          & \multicolumn{1}{c|}{54.975}          & 25.828          \\ \hline
\multirow{2}{*}{\textbf{OSNet IBN}}    & \textbf{IDF1}                    & \multicolumn{1}{c|}{65.769}          & \multicolumn{1}{c|}{65.368}          & 22.269          \\ \cline{2-5} 
                                       & \textbf{MOTA}                    & \multicolumn{1}{c|}{52.892}          & \multicolumn{1}{c|}{55.078}          & 25.140          \\ \hline
\end{tabular}
\end{table}

\subsection{Detection of graph communities with position estimation module}

Our method revolves heavily around the idea of community detection on the graph. The Importance variable is updated every frame for each connection, and based on the strength of the connection's Importance communities are created. We tested the following algorithms for community detection: greedy modularity maximization (GMM) \cite{2004_newman_greedy_mod_det_com}, Louvain (LOU) \cite{2008_louvain}, asynchronous label propagation (ASYN) \cite{2007_asyn}, spectral clustering (SC) \cite{2001_spectral_clusteing} and Girvan–Newman weighted by Importance (GNI) and on betweenness centrality (GNC) \cite{2001_girvan_newman}. In the case of greedy modularity maximization and Louvain methods, we limited the number of possible community members to the number of cameras.

Again, the IDF1 and MOTA were aggregated by median value based on the number of people present in the scene (Supplementary Table 3, Supplementary Figure 3). In Table \ref{tab:communities_idf1_mota}, we present the median, mean, and standard deviation of the results for each community detection method. We found significant differences between the result groups (p-value=0.00237). The post hoc Nemenyi joint with a median of the sets' aggregations revealed that the best-performing method was GMM (relative to LOU p-value=2.125e-08, ASYN p-value=5.596e-14, SC p-value=3.4e-38, GNI p-value=3.4e-38, GNC p-value=3.4e-38). Other methods' median IDF1 scores largely deviate from the median IDF1 of GMM. The Louvain method was promising at first, showing high IDF1 values with a small number of people (2-3 people); however, as the number of people in the space increased, the quality of MOT declined steeply.

\begin{table}[]
\centering
\caption{Community detection algorithms' performance comparison on the internal dataset. The median, mean and standard deviation of the create group metrics are shown in the table.}
\label{tab:communities_idf1_mota}
\begin{tabular}{|c|c|ccc|}
\hline
\multirow{2}{*}{\textbf{\begin{tabular}[c]{@{}c@{}}Community\\ Detection\end{tabular}}} & \multirow{2}{*}{\textbf{Metric}} & \multicolumn{3}{c|}{\textbf{Aggregation}}                                                  \\ \cline{3-5} 
                                                                                        &                                  & \multicolumn{1}{c|}{\textbf{Median}} & \multicolumn{1}{c|}{\textbf{Mean}}   & \textbf{STD} \\ \hline
\multirow{2}{*}{\textbf{GMM}}                                                           & \textbf{IDF1}                    & \multicolumn{1}{c|}{\textbf{79.279}} & \multicolumn{1}{c|}{\textbf{77.608}} & 14.701       \\ \cline{2-5} 
                                                                                        & \textbf{MOTA}                    & \multicolumn{1}{c|}{63.689}          & \multicolumn{1}{c|}{64.202}          & 21.339       \\ \hline
\multirow{2}{*}{\textbf{LOU}}                                                           & \textbf{IDF1}                    & \multicolumn{1}{c|}{39.454}          & \multicolumn{1}{c|}{48.511}          & 26.641       \\ \cline{2-5} 
                                                                                        & \textbf{MOTA}                    & \multicolumn{1}{c|}{36.354}          & \multicolumn{1}{c|}{43.214}          & 24.089       \\ \hline
\multirow{2}{*}{\textbf{ASYN}}                                                          & \textbf{IDF1}                    & \multicolumn{1}{c|}{28.66}           & \multicolumn{1}{c|}{33.597}          & 13.141       \\ \cline{2-5} 
                                                                                        & \textbf{MOTA}                    & \multicolumn{1}{c|}{50.749}          & \multicolumn{1}{c|}{52.730}          & 9.491        \\ \hline
\multirow{2}{*}{\textbf{SC}}                                                            & \textbf{IDF1}                    & \multicolumn{1}{c|}{17.048}          & \multicolumn{1}{c|}{22.269}          & 12.490       \\ \cline{2-5} 
                                                                                        & \textbf{MOTA}                    & \multicolumn{1}{c|}{27.888}          & \multicolumn{1}{c|}{33.402}          & 24.677       \\ \hline
\multirow{2}{*}{\textbf{GNI}}                                                           & \textbf{IDF1}                    & \multicolumn{1}{c|}{29.322}          & \multicolumn{1}{c|}{31.442}          & 12.951       \\ \cline{2-5} 
                                                                                        & \textbf{MOTA}                    & \multicolumn{1}{c|}{80.464}          & \multicolumn{1}{c|}{77.932}          & 10.647       \\ \hline
\multirow{2}{*}{\textbf{GNC}}                                                           & \textbf{IDF1}                    & \multicolumn{1}{c|}{30.063}          & \multicolumn{1}{c|}{32.101}          & 17.067       \\ \cline{2-5} 
                                                                                        & \textbf{MOTA}                    & \multicolumn{1}{c|}{47.374}          & \multicolumn{1}{c|}{48.935}          & 12.549       \\ \hline
\end{tabular}
\end{table}

\subsection{Multi-camera multi-object tracking without position estimation module}

Position estimation is one of the key features of GRAP-MOT, but since it was tailored to an internal dataset, we also evaluated the system without it. When unavailable, a supplementary bounding box relation module estimates positional similarity across cameras. While the feature extraction and community detection modules show clear advantages, the choice of tracking method remains uncertain; hence, additional experiments were conducted with DeepSORT, SORT, and ByteTrack.

We found significant differences in IDF1 (p-value=0.00237) and MOTA (p-value=8.111e-10) metrics between tracking methods (Supplementary Table 4, Supplementary Figure 4, Table \ref{tab:no_distance_idf1_mota} shows the median, mean, and standard deviation of the metrics for recording groups). The post hoc test revealed that this time the DeepSORT was the best-performing method (p-value=4.832e-10 in comparison to SORT; p-value=8.519e-05 in comparison to Byte Track). There was no statistically significant difference between the results of SORT and Byte Track (p-value=0.0693).

\begin{table}[]
\centering
\caption{MOT results without position estimation module; performance comparison on the internal dataset. The median, mean and standard deviation of the created group metrics are shown in the table.}
\label{tab:no_distance_idf1_mota}
\begin{tabular}{|c|c|ccc|}
\hline
\multirow{2}{*}{\textbf{Tracker}}   & \multirow{2}{*}{\textbf{Metric}} & \multicolumn{3}{c|}{\textbf{Aggregation}}                                                  \\ \cline{3-5} 
                                    &                                  & \multicolumn{1}{c|}{\textbf{Median}} & \multicolumn{1}{c|}{\textbf{Mean}}   & \textbf{STD} \\ \hline
\multirow{2}{*}{\textbf{DeepSORT}}  & \textbf{IDF1}                    & \multicolumn{1}{c|}{\textbf{36.076}} & \multicolumn{1}{c|}{\textbf{40.350}} & 14.044       \\ \cline{2-5} 
                                    & \textbf{MOTA}                    & \multicolumn{1}{c|}{11.076}          & \multicolumn{1}{c|}{15.638}          & 12.571       \\ \hline
\multirow{2}{*}{\textbf{SORT}}      & \textbf{IDF1}                    & \multicolumn{1}{c|}{31.984}          & \multicolumn{1}{c|}{34.529}          & 13.435       \\ \cline{2-5} 
                                    & \textbf{MOTA}                    & \multicolumn{1}{c|}{18.238}          & \multicolumn{1}{c|}{19.159}          & 5.585        \\ \hline
\multirow{2}{*}{\textbf{ByteTrack}} & \textbf{IDF1}                    & \multicolumn{1}{c|}{32.991}          & \multicolumn{1}{c|}{36.034}          & 12.470       \\ \cline{2-5} 
                                    & \textbf{MOTA}                    & \multicolumn{1}{c|}{14.116}          & \multicolumn{1}{c|}{16.119}          & 6.771        \\ \hline
\end{tabular}
\end{table}

\subsection{GRAP-MOT evaluation on the external dataset}

We analyzed the CAMPUS dataset to compare our approach with other existing solutions on publicly available real data. Specifically, we used the Auditorium, Garden1, Garden2,, and Parkinglot subsets. Overall, the CAMPUS dataset's tracks are not ideal, as some recordings took place outside, and not all camera views were overlapping. Also, the CAMPUS dataset does not include any information about a person's position; thus, the GRAP-MOT was executed without the position estimation module. We used the best-performing algorithms from the previous experiments (the ResNet50 model and Greedy Modularity Maximisation community detection method) for feature extraction and community detection. The tracking algorithm was chosen based on experiments without the position estimation module (DeepSORT tracking method).

It is impossible to compare IDF1 with other methods because the authors did not provide them in their articles. In terms of MOTA, the GRAP-MOT method is comparable to TRACTA, STP, HCT, KSP, and POM aside from the Garden1 results (Table \ref{tab:external_dataset}). In comparison with the DyGLIP, the MOTA metric differences are high. However, this result only means that there were not many ID switches, since MOTA does not supply information about the wellness of the label assignment.

\begin{table}[]
\centering
\caption{MOT results without position estimation module; performance comparison on the CAMPUS external dataset. Results are given mainly for the MOTA parameter, which is only available for other methods.}
\label{tab:external_dataset}
\scalebox{0.9}{
    \begin{tabular}{|c|c|cccc|}
    \hline
    \multirow{2}{*}{\textbf{Method}}   & \multirow{2}{*}{\textbf{Metric}} & \multicolumn{4}{c|}{\textbf{Recording}}                                                                                                        \\ \cline{3-6} 
                                       &                                  & \multicolumn{1}{c|}{\textbf{Auditorium}} & \multicolumn{1}{c|}{\textbf{Garden1}} & \multicolumn{1}{c|}{\textbf{Garden2}} & \textbf{Parkinglot} \\ \hline
    \multirow{2}{*}{\textbf{GRAP-MOT}} & \textbf{IDF1}                    & \multicolumn{1}{c|}{26.69}               & \multicolumn{1}{c|}{14.94}            & \multicolumn{1}{c|}{19.02}            & 24.55               \\ \cline{2-6} 
                                       & \textbf{MOTA}                    & \multicolumn{1}{c|}{22.98}               & \multicolumn{1}{c|}{18.68}            & \multicolumn{1}{c|}{35.29}            & 33.15               \\ \hline
    \multirow{2}{*}{\textbf{DyGLIP}}   & \textbf{IDF1}                    & \multicolumn{1}{c|}{-}                   & \multicolumn{1}{c|}{-}                & \multicolumn{1}{c|}{-}                & -                   \\ \cline{2-6} 
                                       & \textbf{MOTA}                    & \multicolumn{1}{c|}{96.7}                & \multicolumn{1}{c|}{71.2}             & \multicolumn{1}{c|}{87}               & 72.8                \\ \hline
    \multirow{2}{*}{\textbf{TRACTA}}   & \textbf{IDF1}                    & \multicolumn{1}{c|}{-}                   & \multicolumn{1}{c|}{-}                & \multicolumn{1}{c|}{-}                & -                   \\ \cline{2-6} 
                                       & \textbf{MOTA}                    & \multicolumn{1}{c|}{33.7}                & \multicolumn{1}{c|}{58.5}             & \multicolumn{1}{c|}{35.5}             & 39.4                \\ \hline
    \multirow{2}{*}{\textbf{STP}}      & \textbf{IDF1}                    & \multicolumn{1}{c|}{-}                   & \multicolumn{1}{c|}{-}                & \multicolumn{1}{c|}{-}                & -                   \\ \cline{2-6} 
                                       & \textbf{MOTA}                    & \multicolumn{1}{c|}{24}                  & \multicolumn{1}{c|}{57}               & \multicolumn{1}{c|}{30}               & 28                  \\ \hline
    \multirow{2}{*}{\textbf{HCT}}      & \textbf{IDF1}                    & \multicolumn{1}{c|}{-}                   & \multicolumn{1}{c|}{-}                & \multicolumn{1}{c|}{-}                & -                   \\ \cline{2-6} 
                                       & \textbf{MOTA}                    & \multicolumn{1}{c|}{20.6}                & \multicolumn{1}{c|}{49}               & \multicolumn{1}{c|}{25.8}             & 24.1                \\ \hline
    \multirow{2}{*}{\textbf{KSP}}      & \textbf{IDF1}                    & \multicolumn{1}{c|}{-}                   & \multicolumn{1}{c|}{-}                & \multicolumn{1}{c|}{-}                & -                   \\ \cline{2-6} 
                                       & \textbf{MOTA}                    & \multicolumn{1}{c|}{17.6}                & \multicolumn{1}{c|}{28.1}             & \multicolumn{1}{c|}{21.9}             & 14                  \\ \hline
    \multirow{2}{*}{\textbf{POM}}      & \textbf{IDF1}                    & \multicolumn{1}{c|}{-}                   & \multicolumn{1}{c|}{-}                & \multicolumn{1}{c|}{-}                & -                   \\ \cline{2-6} 
                                       & \textbf{MOTA}                    & \multicolumn{1}{c|}{16.2}                & \multicolumn{1}{c|}{22.4}             & \multicolumn{1}{c|}{14}               & 11                  \\ \hline
    \end{tabular}
    }
\end{table}

Additionally, we compared our method with the ReST model \cite{2023_rest}. To get IDF1 and MOTA, we used the existing implementation of the ReST method shared on GitHub. Each recording from the CAMPUS dataset was evaluated with its respective spatial and temporal graphs. Tests were conducted based on the last 20\% of recordings' frames, as the first 80\% was used for graphs training. Our method, GRAP-MOT, was evaluated using the same number of frames to match the experimental setup.

\begin{table}[h]
\centering
\caption{MOT results made on the frames used for ReST method evaluation; Garden1: 2280-2849, Garden2: 4800-6000 and Parkinglot: 5828-6475. It is not possible to compare methods on whole recordings because the temporal and spatial graphs of ReST method were trained on previous frames.}
\label{tab:last_20perc_our_vs_rest}
\begin{tabular}{|c|c|ccc|}
\hline
\multirow{2}{*}{\textbf{Method}}   & \multirow{2}{*}{\textbf{Metric}} & \multicolumn{3}{c|}{\textbf{Recording}}                                                             \\ \cline{3-5} 
                                   &                                  & \multicolumn{1}{c|}{\textbf{Garden1}} & \multicolumn{1}{c|}{\textbf{Garden2}} & \textbf{Parkinglot} \\ \hline
\multirow{2}{*}{\textbf{GRAP-MOT}} & \textbf{IDF1}                    & \multicolumn{1}{c|}{39.84}            & \multicolumn{1}{c|}{33.69}            & 38.16               \\ \cline{2-5} 
                                   & \textbf{MOTA}                    & \multicolumn{1}{c|}{33.11}            & \multicolumn{1}{c|}{33.61}            & 23.26               \\ \hline
\multirow{2}{*}{\textbf{ReST}}     & \textbf{IDF1}                    & \multicolumn{1}{c|}{27.3}             & \multicolumn{1}{c|}{32.2}             & 25.0                \\ \cline{2-5} 
                                   & \textbf{MOTA}                    & \multicolumn{1}{c|}{78.5}             & \multicolumn{1}{c|}{85.5}             & 76.7                \\ \hline
\end{tabular}
\end{table}

The Auditorium recording was not tested because the authors did not supplement the spatial and temporal graphs of the ReST method. The GRAP-MOT scored best in terms of IDF1 on each recording (Table \ref{tab:last_20perc_our_vs_rest}). For Garden1 and Parkinglot the differences in IDF1 values are 12.54 and 13.16 respectively; for Garden2 the difference is 1.49. The ReST method scored best in terms of MOTA, with differences in values being for Garden1 38.66, Garden2 51.86, and 53.44. The reason for such large discrepancies between the IDF1 and MOTA values of the two methods is explained in the discussion, while also showing examples of why IDF1 is a better measure for assessing MOT quality.

\begin{figure}[h]
    \centering
    \includegraphics[width=1\linewidth]{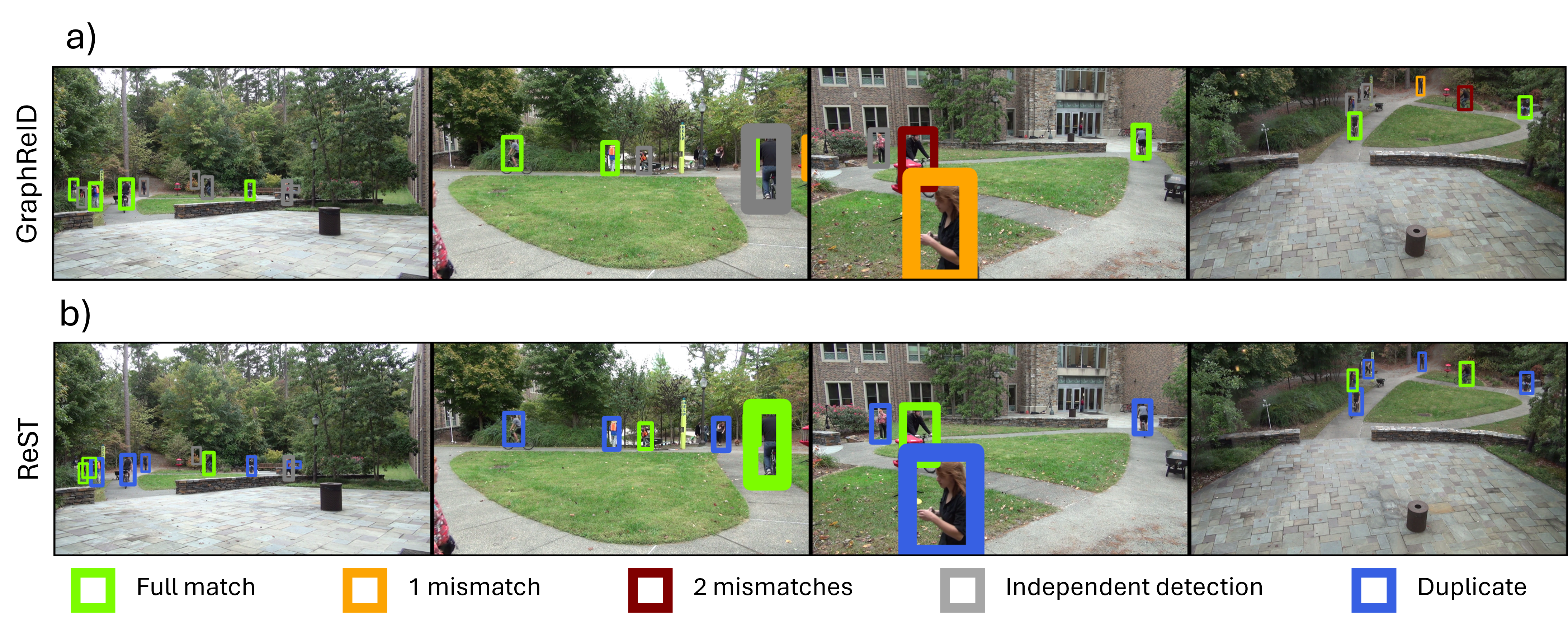}
    \caption{Example frame from the Garden2 recording illustrating discrepancies between IDF1 and MOTA. (a) GRAP-MOT final pipeline with a few ID mismatches, yielding IDF1 = 38.66 and MOTA = 48.09. (b) ReST with spatial and temporal graphs trained on Garden2 shows low IDF1 = 32.2 but high MOTA = 85.5 due to duplicated IDs. Color coding: green – correctly tracked detections across cameras; orange – mismatch on one camera; red – mismatch on two cameras; blue – repeated ID on one camera; grey – detections missing in other frames.}
    \label{fig:our_vs_rest}
\end{figure}

\section{Discussion}
We proposed a novel method for detecting and tracking people in closed spaces that effectively combines information from multiple video cameras. We tested several solutions for tracking, feature extraction, and community detection on our internal dataset and highlighted the best algorithms: SORT, ResNet50 model, and Greedy Modularity Maximisation community detection method. Then, we used the CAMPUS dataset to compare the GRAP-MOT with other state-of-the-art methods. The requirement of a highly congested space with overlapping camera views was satisfied by the Garden1, Garden2, and Parkinglot subsets, and the requirement of closed space was met by the Auditorium and Parkinglot subsets. We obtained comparable MOTA metric values to the TRACTA method and better than STP, HCT, KSP, and POM (Table \ref{tab:external_dataset}). The DyGLIP reports outstanding results of the MOTA parameter, but as we describe later, it's not the best metric. From existing algorithms, we managed to run only the ReST method (Table \ref{tab:last_20perc_our_vs_rest}). Large gap between the IDF1 and MOTA when using  ReST results mostly from duplicate detections in the scope of one recording. At times when the ReST model can not decide which neighboring detections should have a specific ID, it assigns the same ID to both detections. This eliminates the ID switches, resulting in a larger MOTA, but lower IDF1. An example frame from the Garden2 recording is shown in Figure \ref{fig:our_vs_rest} where the top panel contains GRAP-MOT matches and the bottom panel ReST matches. We did not use the Market1501 \cite{2015_market1501}, Mars \cite{2016_mars}, or CUHK03 \cite{2014_cuhk03} datasets for evaluation since they do not meet the above-specified requirements.

When calculating the Importance metric \ref{eq_importance_definition}, image bounding box features and position estimation were the base for the MOT anchor, with occurrences solidifying the connection between vertices. When the position estimation module is off \ref{"eq_new_importance_after_rep"}, only features are a solid base for the MOT anchor; the $r$ bounding box relation coefficient is a rough position estimation meant to strengthen the already existing anchor. This makes the GRAP-MOT approach with position estimation solidify connections between proper vertices much faster because both features and position work together. Without position estimation, the GRAP-MOT first analyzes features, and only after some time checks how bounding boxes are located in relation to each other. The position estimation plays an important role in connecting all the MOT elements; thus, the SORT tracking method reports the best results (Table \ref{tab:tracking_idf1_mota}). On the other hand, without the position data, deeper information is required, and DeepSORT gives better results (Table \ref{tab:no_distance_idf1_mota}). But still, the difference in medians between the GRAP-MOT with and without the position estimation module for IDF1 is 81.492 to 36.076, and for MOTA 72.269 to 11.076, respectively. Taking those results into consideration, the position data should improve results for external datasets as well.


Our problem revolves mainly around human multi-camera multi-object tracking. While sharing many similarities with more popular vehicle tracking, it covers a much wider variety of backgrounds and has to deal with occlusion, viewpoint variations, and pose variations. In the person MOT task, aside from the bounding box features, additional information could be used, like trajectory \cite{2022_cars_mc_mt_reId}, homography matrix information to map detections across cameras \cite{2023_rest}, pose \cite{2018_pose_1, 2018_pose_2} or specific body part features \cite{2023_body_parts}. However, in the environment presented in our dataset, the use of other body parts features, people's poses, and their walking trajectories is not possible. To supplement this, we demonstrate the effectiveness of the exact position determination method, while also proposing an alternative approach that replaces the use of the homography matrix, acknowledging that it may not always be available. Applying various deep learning methods is the most popular approach \cite{2021_centroids_reid, 2020_unsupervised_reid, 2019_temporal_reid, 2023_rest} since their ability to gather features from the images is unrivaled. However, the availability of working implementations for person MOT is very limited and, if they exist, it is very difficult to cope with the complexities of the programming environment they impose. Although code for methods such as TRACTA and DyGLIP is available on the GitHub platform, we have not been able to run them. In the case of TRACTA, this was due to requirements such as the CAFFE library, which ceased to be supported in 2014, and its installation requires compiling it from source and installing a suitably old version of Linux under that. In the case of the DyGLIP method, despite our sincere intentions, we noticed that a key script was missing. Without it, the second step (graph features extraction) of the method is impossible to execute. The authors are aware of the absence of this piece of code. In the case of ReST, we managed to run the provided code without major problems.

\begin{figure}[h]
    \centering
    \includegraphics[width=0.99\linewidth]{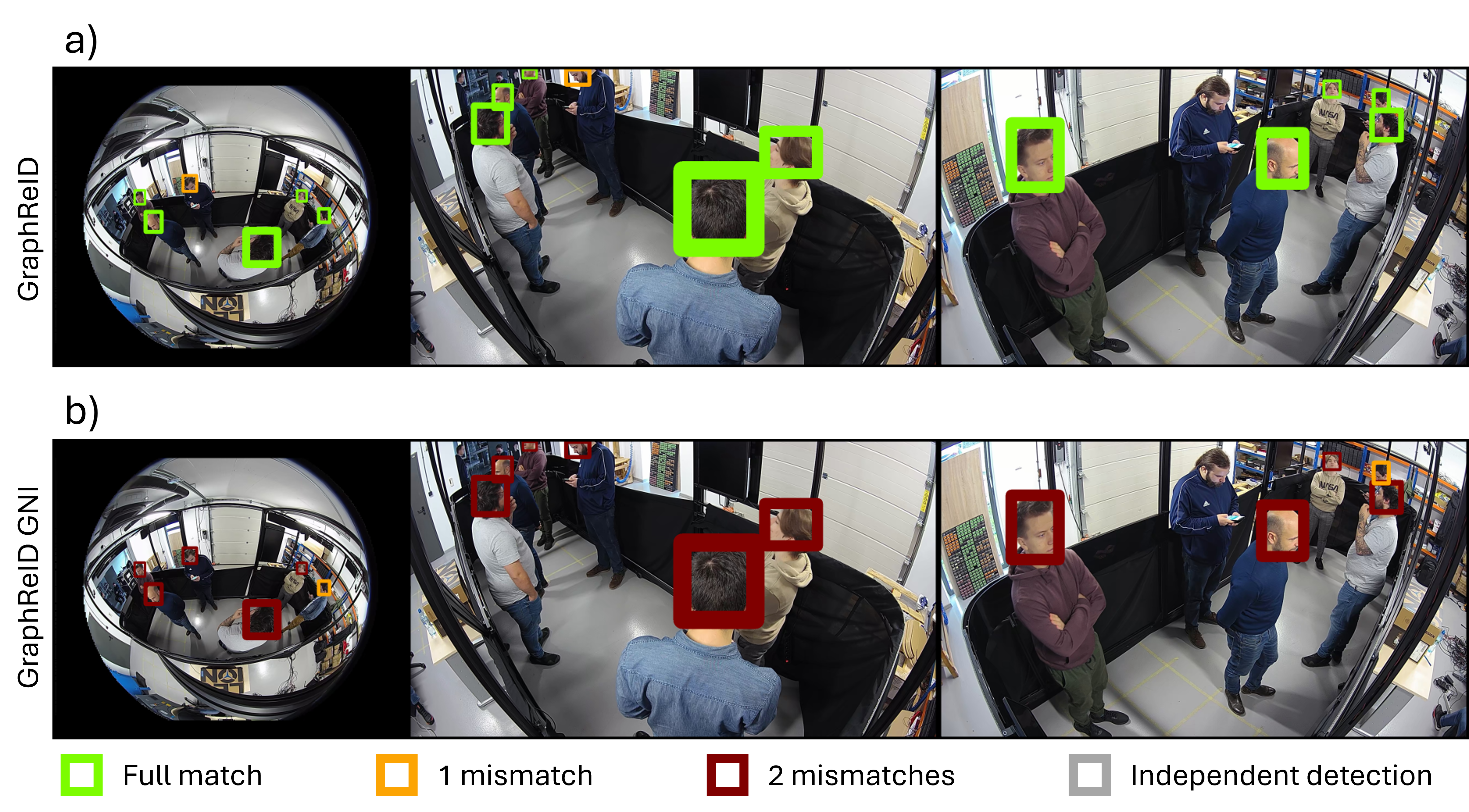}
    \caption{Example frame from the internal gt6\_task10 recording highlighting the mismatch between IDF1 and MOTA. a) GRAP-MOT final pipeline, where both IDF1=97.607 and MOTA=95.214 indicate relatively correct results, b) GRAP-MOT testing GNI community detection algorithm for which mismatches are visible in IDF1=39.567 while MOTA=86.463 is still very high. Green: all detections tracked correctly across the multi-camera views, orange: MOT mismatch on one camera, red: MOT mismatch on two cameras and grey: detections not tracked on other frames.}
    \label{fig:gni_idf1_vs_mota}
\end{figure}

After careful analysis, we consider IDF1 rather than MOTA the most important metric for person MOT performance testing (but we reported both parameters in our results). In the MOT task, ensuring the correct and consistent assignment of identifiers to detections across multiple cameras is critical for preserving identity tracking. MOTA penalizes the method primarily for errors such as missed detections and false alarms (both directly tied to the presence or absence of the bounding boxes), which means that even incorrectly assigned identifiers could still give high MOTA values. Also, identifier switches are penalized, but a direct measure of the impact of this change on identifier correctness is not provided. In contrast, IDF1 specifically measures the quality of identifier assignments and consistency across all frames. Therefore, we believe that IDF1 is a better scoring metric for MOT tasks than MOTA. There are situations when a high discrepancy between both indices is observed (Supplementary Figure 5). An example of where IDF1 is a better MOT quality measure than MOTA is an experiment concerning community detection methods (Table \ref{tab:communities_idf1_mota}). When analyzing the result of the experiment in terms of MOTA values, the clear winner is GNI (Girvan–Newman weighted by Importance). Post hoc analysis indicates statistical significance over other methods, with values ranging from p-value=8.389e-07, in the case of GMM, up to p-value=3.4e-38, in the case of SC. This results from stagnation, i.e. about halfway through the recording method stops changing the IDs of the tracklets. This results in no ID switches and a constant rise in MOTA. The label assignment is still wrong, which is neglected by the MOTA metric (Figure \ref{fig:gni_idf1_vs_mota}).

In a closed space, the main cause of tracklet interference is the presence of obstructions. When an obstruction occurs, a node linked to a disappearing detection stops receiving updates with new image feature values. However, the node itself does not disappear; its stored values remain in memory for a period of time. Community formation continues based on this older information. If a connection is formed during this time, the occurrences value is also updated, which strengthens the connection. This process helps maintain the continuity of the tracklet despite temporary interruptions. To test the robustness of the proposed method against occlusions, we designed an experiment in which occlusions were simulated by removing random detections for 20 consecutive frames every 20 frames from one, two, or three cameras. The tests were performed on 10 recordings from the scene named gt10 and compared to the reference recordings (with all detections present) using a T-test (Table \ref{table:occlusion}). The T-test shows that there is no statistically significant difference between the IDF1 and MOTA values of recordings with and without occlusions, assuming a significance level $\alpha=0.05$.

\begin{table}[h]
    \caption{Median IDF1 and MOTA of the gt10 scene recordings, along with T-test p-values comparing the recordings with occlusions simulated on one, two, and three cameras to the reference recordings without occlusions.}
    \label{table:occlusion}
    \centering
    \begin{tabular}{|c|c|c||c|c|}
        \hline
        \textbf{Cameras}           & \textbf{IDF1} & \textbf{p-value} & \textbf{MOTA} & \textbf{p-value} \\ \hline
        \textbf{-}                 & 71.057        & -                & 53.178        & -                \\ \hline
        \textbf{fe}                & 71.057        & 1.0              & 53.178        & 1.0              \\ \hline
        \textbf{fe + left}         & 70.648        & 0.927            & 52.826        & 0.869            \\ \hline
        \textbf{fe + left + right} & 70.410        & 0.825            & 52.708        & 0.863            \\ \hline
    \end{tabular}
\end{table}

\mycomment{
    \begin{figure}[h]
        \centering
        \includegraphics[width=0.99\linewidth]{Figures/Oclussions.png}
        \caption{Example 75th, 88th and 145th frame, showing first appearance of detection, first loss of detection, and reappearance of the detection on the camera shown in the middle subplot.}
        \label{fig:occlusion}
    \end{figure}
}

GRAP-MOT has several limitations. It is designed for short recordings in enclosed spaces, such as rooms, corridors, warehouses, or public transport, where people are densely packed, partially occluded, and visible from multiple camera angles. In this study, we applied head detection instead of full-body detection to improve visibility and resistance to occlusion. New tracklets are initially uncertain, and their IDs may fluctuate during the first few frames. The method assumes that people within the field of view rarely leave it, and stabilization of tracklet IDs occurs only after the Importance value stabilizes. This delayed stabilization can lower MOTA scores compared to other methods, as seen in the Garden1 recording. Garden1 footage is outdoors, showing a large group of distant individuals who are close together in one camera view. At this distance, feature extraction provides limited information, and proximity-based analysis fails to separate tracklets effectively. Additionally, the frequent entry and exit of people in the early frames prevent the algorithm from achieving stabilization in time.

Future work aims to adapt the system for open spaces without overlapping camera views. Extending the lifespan of graph edges and refining feature management could improve performance in such environments. Network and code optimizations may reduce computational time. The current network is designed for full-body analysis and trained on Market1501 and DukeMTMC datasets, limiting its suitability for the present problem. Retraining on datasets like Wildtrack \cite{2018_wildtrack} could enhance efficiency.

\section{Conclusions}
In the presented work, we developed a new method for tracking multiple persons under conditions of observation from many cameras in a closed space. We tested the proposed algorithm on publicly available data and on data that we recorded ourselves. We also conducted an in-depth analysis of comparative parameters used to analyze the MOT problem. Finally, we put great emphasis on data access and the transparency, and ease of running our source code.

\bibliographystyle{IEEEtran}

\end{document}